\documentclass[twocolumn]{article}
\usepackage{spconf,amsmath,epsfig}
 \usepackage{booktabs}
 \usepackage{graphicx}
\usepackage{colortbl}
\usepackage{subcaption}
\usepackage{changes}
\title{Mitigating Bad Ground Truth in Supervised Machine Learning based Crop Classification: A Multi-Level Framework with Sentinel-2 Images}
\name{\begin{tabular}{c}
       Sanayya A, Amoolya Shetty, Abhijeet Sharma,\\
       Venkatesh Ravichandran, Masthan Wali Gosuvarapalli,\\
       Sarthak Jain, Priyamvada Nanjundiah, Ujjal Kr Dutta, Divya Sharma
       \end{tabular}}
\address{SatSure Analytics India Pvt Ltd}

\begin{document}
%\ninept
%

\maketitle
\begin{abstract}
In agricultural management, precise Ground Truth (GT) data is crucial for accurate Machine Learning (ML) based crop classification. Yet, issues like crop mislabeling and incorrect land identification are common. We propose a multi-level GT cleaning framework while utilizing multi-temporal Sentinel-2 data to address these issues. Specifically, this framework utilizes generating embeddings for farmland, clustering similar crop profiles, and identification of outliers indicating GT errors. We validated clusters with False Colour Composite (FCC) checks and used distance-based metrics to scale and automate this verification process. The importance of cleaning the GT data became apparent when the models were trained on the clean and unclean data.  For instance, when we trained a Random Forest model with the clean GT data, we achieved upto 70\% absolute percentage points higher for the F1 score metric. This approach advances crop classification methodologies, with potential for applications towards improving loan underwriting and agricultural decision-making.
\end{abstract}
\begin{keywords}
Ground Truth, Quality Assessment, Sentinel-2, Crop Classification, Deep learning, Machine learning
\end{keywords}
\vspace{-0.4cm}
\section{Introduction}
\label{sec:intro}
Agricultural lending strategies frequently hinge on accurate crop identification, which is vital for effective loan credit underwriting \cite{ghosh2008problems,benami2021uniting}. The process of crop classification, however, is complicated by the variability in cropping patterns, agricultural practices, and other influencing factors \cite{lin2022early,kumar2017statistical}. An alternative to manual verification is to automate the process of crop classification to aid the existing underwriting processing while potentially removing errors inherent to human data collection at scale. Given the extensive scale at which precise crop information is required—particularly in an agriculture-dependent nation like India—implementing automated crop classification systems is imperative.  Machine learning presents a robust framework for automating this process, enhancing the efficiency of crop identification by leveraging remote sensing data and reducing dependency on extensive in-person ground verification. Several studies have looked at the issue of crop classification from the lens of varied datasets in the recent past \cite{kumar2017statistical}. Nonetheless, the efficacy of most machine learning models is contingent on high-quality Ground Truth (GT) data, which is traditionally obtained through manual field verification. This method is impractical on the vast geographical scale of India. 
This study proposes a comprehensive, multi-level framework for cleaning the manually collected GT data, and utilising the same to classify three major Rabi crops (mustard, paddy, wheat) using multi-temporal optical data based Machine Learning (ML). By leveraging large-scale crop classification to support informed agricultural decision-making, we aim to enhance loan collection efficiency and credit underwriting.
\section{Proposed Multi-Level Framework}
To account for the diversity in cropping in a country like India, we utilise data collected across various agro-climatic regions and multiple growing seasons to capture sufficient spatial and temporal variation. Accurate ground truth (GT) data is fundamental to reliable crop classification in agricultural management. However, GT data collected by field vendors are subject to errors inherent to manual data collection, including the mislabeling of crops and the incorrect identification of the farmland \cite{joshi2023remote}. These inaccuracies can lead to flawed crop classification, affecting agricultural policy decisions and loan underwriting processes.\\
\begin{figure}[!tb]
\centering
\includegraphics[width=8.5cm]{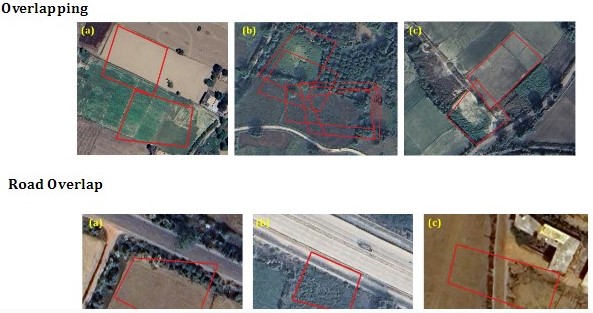}
\caption{\textbf{Level 1 (L1)}: Overall processing and mislabeling, focusing on plots mapped to non-agricultural use cases, excessive overlap of multiple ground truth (GT) polygons, and overlaps with roads or built structures}
\label{fig:level_one_processing}
\end{figure}
\vspace{-0.5cm}
 \subsection{Multi-Level Elimination}
\begin{figure*}[!tb]
\centering
\includegraphics[width=0.8\textwidth]{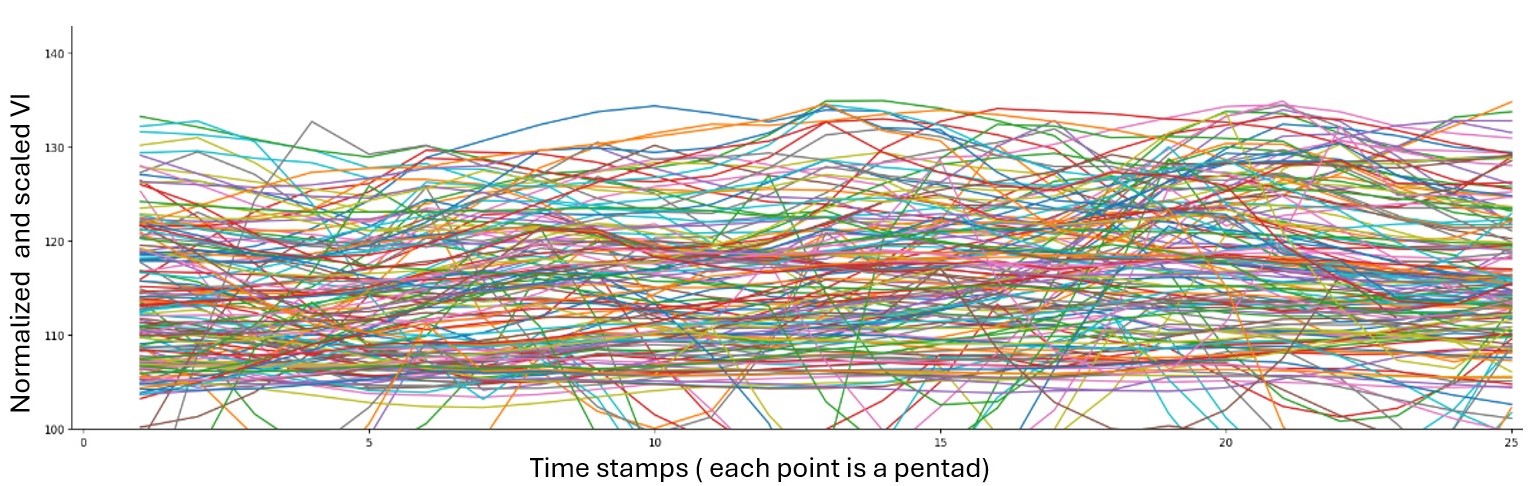}
\caption{\textbf{Level 2 (L2)}: Cleaning GT data using VI temporal profiles. The sheer intertwining of profiles seen here is untangled by isolating those profiles that fall below a prescribed VI value.}
\label{fig:level_two_cleaning}
\end{figure*}

The process is divided into four levels to obtain the most optimal GT points. Each level focuses on different aspects of errors present during manual GT collection. In \textbf{Level 1 (L1)}, all the plots that fell on roads, intersections and, in general, non-agricultural areas were eliminated by manual inspection of the collected polygons with  Google Earth base maps. The focus here was also to eliminate those polygons that fell over multiple crops, as that would lead to misleading signals when training a ML model [Fig. \ref{fig:level_one_processing}].
\textbf{Level 2 (L2)} of the GT cleaning process was a statistical approach carried over the polygons that cleared level 1. We used multi-temporal Sentinel 2 data to create an NDVI profile for all the pixels within a polygon. All profiles whose maximum NDVI values fell below the threshold generally observed for crops and cultivated vegetation were eliminated here. This step provided the dual benefits of eliminating mislabeled fields that might have been visually interpreted as cultivated plots and eliminating any boundary or non-crop pixels embedded within a field otherwise of superior quality [Fig.  \ref{fig:level_two_cleaning}].
\vspace{-0.4cm}
\subsection{Spectral profile Analysis}

\begin{figure*}[!tb]
\centering
\begin{subfigure}{\textwidth}
    \centering
    \includegraphics[width=0.6\textwidth]{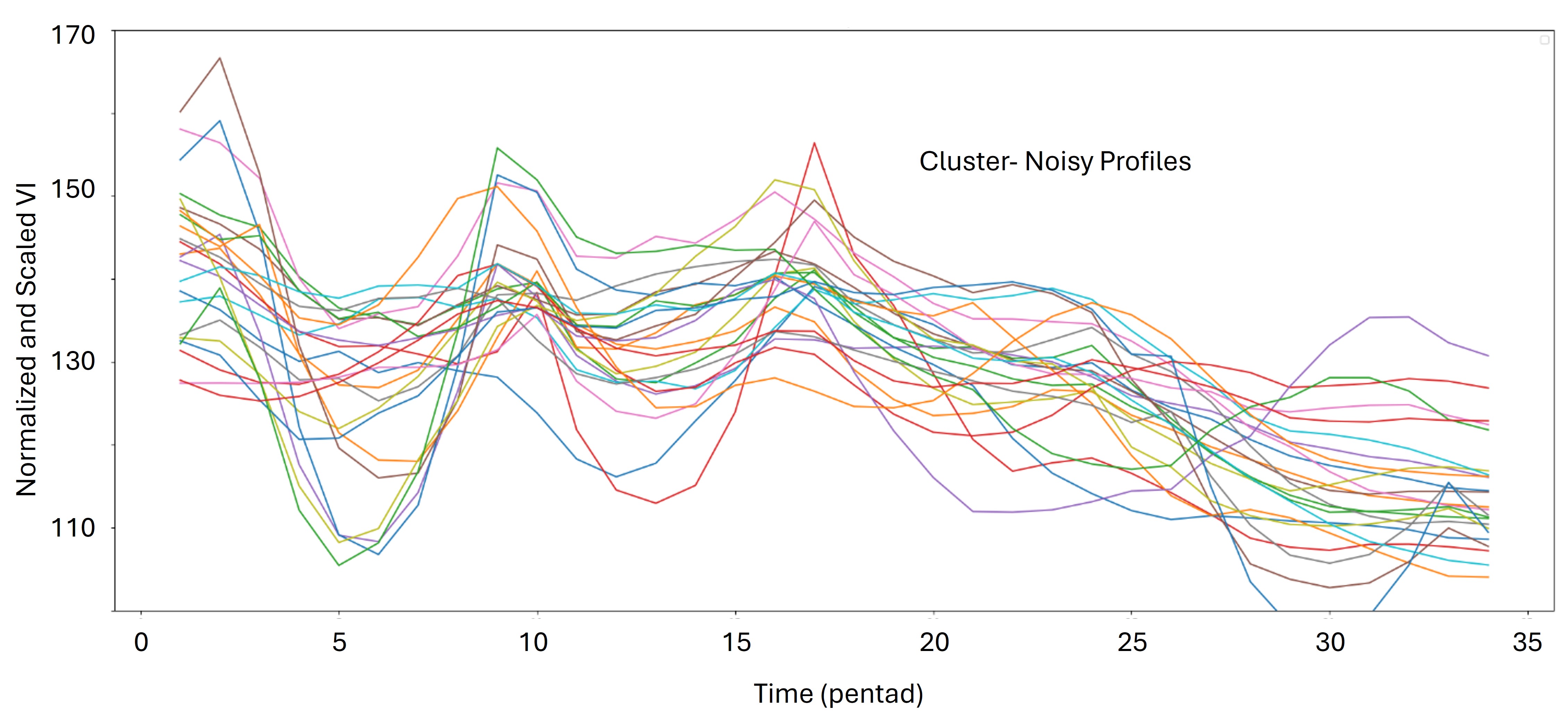}
    \caption{Cluster highlighting noisy profiles, they could be spanning multiple fields or have other issues}
    \label{fig:sub_first_image}
\end{subfigure}\hfill
\begin{subfigure}{\textwidth}
    \centering
    \includegraphics[width=0.6\textwidth]{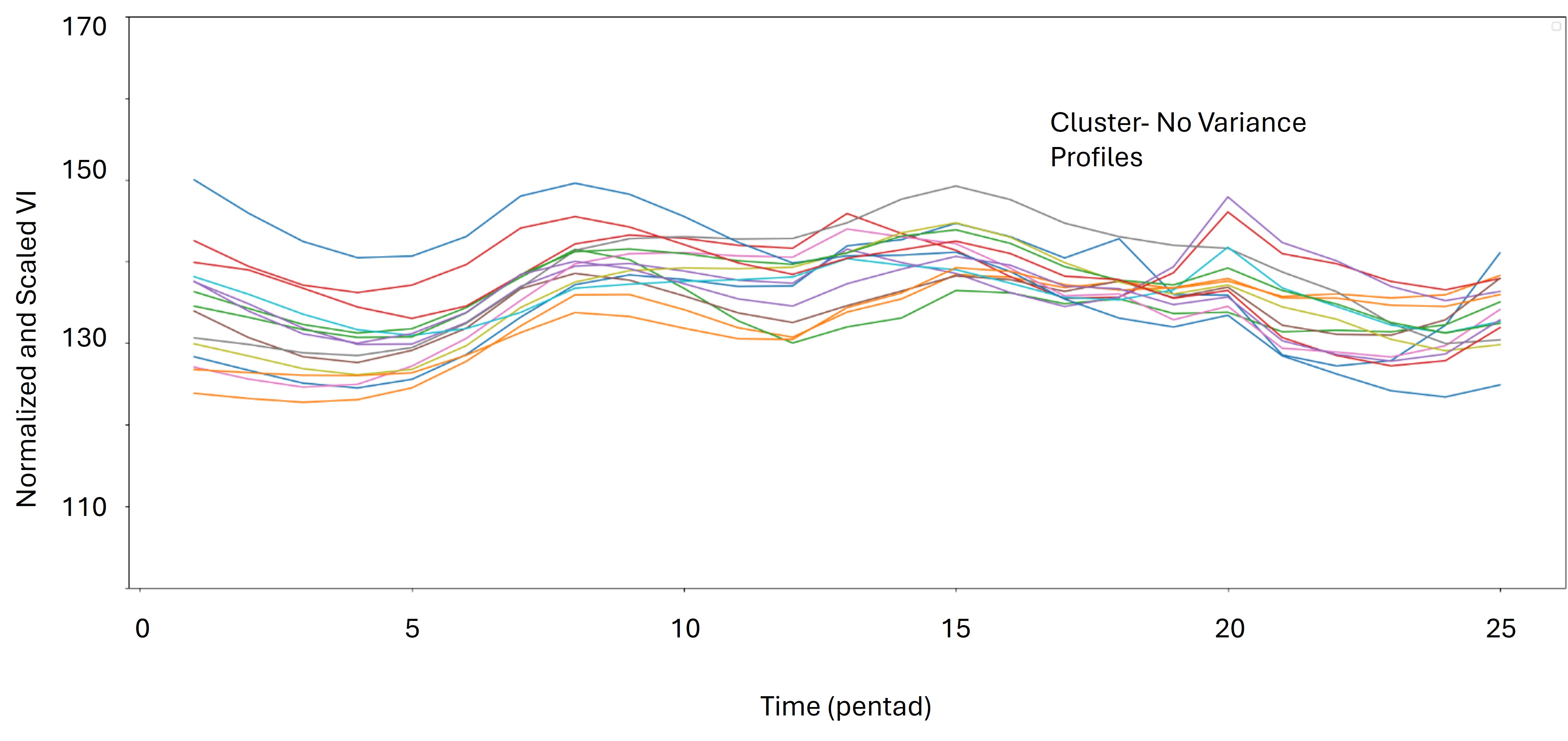}
    \caption{Profiles with very low variability, these are most likely those profiles where no crops were grown}
    \label{fig:sub_second_image}
\end{subfigure}
\caption{\textbf{Level 3 (L3)}: Clustering profiles highlight profiles with low variance as well as noisy profiles.}
\label{fig:level_three_processing}
\end{figure*}

Following level 2 was \textbf{Level 3 (L3)} of cleaning, a clustering-based approach. We used K-means clustering to group the various NDVI temporal profiles to (a) group like signals together and (b) to isolate noisy signals to more easily eliminate them. This process also clustered groups of profiles with very low variance that most likely corresponded to long-term vegetation, which was beyond the scope of the current study. Thus, we have so far eliminated non-agricultural fields, mislabeled plots, non-crop pixels and also those cultivated regions that fall beyond the scope of the crops in question [Fig.  \ref{fig:level_three_processing}]. 

So far, we have focused on qualitative and statistical errors that might have been introduced into the data; we now move into a more nuanced assessment of the quality of the label. While clustering NDVI profiles that are more or less similar, typically, clustering algorithms like K-means are less sensitive to minor variations in the overall shape of the curve. This lack of sensitivity could prove detrimental when we want to classify the collected plots further into their respective crop types. Thus to go a step further, we isolate pixel-wise temporal profiles of each of the following bands of Sentinel-2 satellite data, with a focus on five critical spectral bands: red, green, blue, near-infrared (NIR), and short-wave infrared (SWIR2)\cite{kumar2017statistical,joshi2023remote}. We could capture each crop's unique temporal and spectral signature by calculating spectral embeddings for farmland areas where GT data collection was conducted. This detailed spectral representation provided a comprehensive understanding of each farmland's characteristics, enabling us to distinguish between different crops and identify anomalies.
We then conducted a proprietary False Colour Composite (FCC) analysis on a selection of farms within each cluster. This visual inspection allowed us to confirm the accuracy of the GT labels and correct any discrepancies. While manual inspection is the most robust technique so far, it is by far the most time-consuming part of the process and is not feasible for large areas.

% \begin{equation}
%      \cos(\theta) = \frac{\mathbf{a} \cdot \mathbf{b}}{\|\mathbf{a}\| \cdot \|\mathbf{b}\|}
% \end{equation}

We automated the identification and validation of crop plots in nearby districts to scale the FCC-based verification process. Initially, experts manually verified that selected plots were correctly labelled. We calculated a median spectral profile for each crop type using these verified plots. With these median profiles as benchmarks, we employed distance metrics such as cosine similarity, Pearson correlation, and Manhattan distance \cite{scikit-learn} to group similar crop profiles across neighbouring districts. This approach enabled efficient and accurate GT data verification over larger geographic areas. By adopting these advanced methodologies, we significantly improved the quality and reliability of our GT data, leading to enhanced performance of our crop classification models.
 
 We carry out other pre-processing steps while cleaning the collected ground truth data. Eliminating cloudy pixels, interpolation, and smoothing of the profiles enhance the features for classification and outlier detection to identify the optimal features better and reduce the chances for misclassification, thus ensuring high data quality.

 \section{ML Model evaluation with Newly cleaned data}
Following comprehensive GT cleaning, we evaluated a range of ML models to assess their performance, namely, LightGBM \cite{lightgbm}, Support Vector Machine (SVM) and  Random Forest \cite{joshi2023remote}. Among these, the Random Forest model demonstrated superior performance, achieving impressive F1 scores of 92\% for mustard, 97\% for paddy, and 94\% for wheat when trained on the cleaned ground truth (GT) data (Table \ref{tab:my-table}). \textbf{In stark contrast, the same Random Forest model trained on the uncleaned GT data produced significantly lower F1 scores}: 26\% for mustard, 53\% for paddy, and 25\% for wheat.

The performance improvement is most evident when comparing the cleaned and uncleaned GT data results, as illustrated in Table \ref{tab:performance_metrics}. For instance, the F1 score for mustard in districts like Chhatarpur surged from a mere 10\% to nearly 97\% following data cleaning. \textbf{The most pronounced improvements are observed between the second (L2) and third (L3) levels of processing}. \textit{Third-level (L3) processing, which involves clustering-based cleaning, proved particularly effective}. This method enhanced the model’s accuracy by removing non-crop profiles and significantly refined the overall classification results by eliminating excessively noisy profiles, as well as those that showed very little variability, most likely due to them being non-crop plots.
%\vspace{-1cm}
    \section{Results}
  \vspace{-0.4cm}
The rigorous approach we adopted bolstered the accuracy of crop identification and provided a stronger foundation for agricultural policy development and loan decision-making. The multi-level cleaning process for the GT data effectively addressed various types of errors. [Table \ref{tab:performance_metrics}] illustrates the impact of cleaning procedures on the quality of GT data. Removing mislabeled and non-agricultural points, statistical and clustering-based refinement led to a substantial reduction in erroneous data points. This was critical in ensuring the high quality of input data for subsequent crop classification. As a result, our work has led to meaningful advancements in agricultural management and policy-making, underpinned by the precise and trustworthy GT data obtained through this comprehensive process.
The methodologies applied in this study demonstrated their scalability and generalisability. By extending the accuracy of GT verification across larger geographic areas, we ensured that the improved classification performance is not limited to specific regions but applicable to broader contexts. This approach has potential implications for large-scale agricultural monitoring and policy-making.
\vspace{-0.3cm}
\section{Conclusion}
\textbf{In conclusion}, our results highlight the effectiveness of the proposed GT cleaning and crop classification techniques. The advancements achieved through this comprehensive process have set a new standard for high-quality agricultural data and model performance.

\begin{table}[!htbh]
\centering
\begin{tabular}{lccc}
\hline
& \multicolumn{3}{c}{Unclean GT} \\ \cline{2-4}
Crop    & Recall & Precision & F1-Score \\ \hline
Paddy   & 60   & 47      & 53     \\
Mustard & 23   & 29      & 26     \\
Wheat   & 24   & 26      & 25     \\ \hline
& \multicolumn{3}{c}{Cleaned GT} \\ \cline{2-4}
Crop    & Recall & Precision & F1-Score \\ \hline
Paddy   & 97   & 98      & 97     \\
Mustard & 86   & 99      & 92     \\
Wheat   & 99   & 89      & 94     \\ \hline
\end{tabular}
\caption{Performance metrics (in \%) of Random Forest model for different crops trained on unclean and cleaned GT from 2022-23}
\label{tab:my-table}
\end{table}
 %\section{Discussion}

\begin{table}[!tbh]
\definecolor{lightgray}{gray}{0.8}
\centering
\resizebox{0.5\textwidth}{!}{%
\begin{tabular}{lcccccc}
%\toprule
\rowcolor{lightgray}
\textbf{District} & \textbf{Crop} & \textbf{Year} & \textbf{Level} & \textbf{Precision} & \textbf{TPR} & \textbf{F1} \\ \hline
Bhind(MP) & WH & 2024 & L1 & 58 & 66 & 62 \\ 
          &    &     & L2 & 59 & 67 & 62 \\ 
          \rowcolor{lightgray}
          &    &     & L3 & 99 & 99 & 99 \\ \hline
Bhind(MP) & MU & 2024 & L1 & 70 & 53 & 61 \\ 
          &    &     & L2 & 70 & 54 & 61 \\ 
          \rowcolor{lightgray}
          &    &     & L3 & 100 & 84 & 91 \\ \hline
Ch-pur(MP) & WH & 2024 & L1 & 10 & 49 & 17 \\ 
           &    &     & L2 & 10 & 52 & 17 \\ 
           \rowcolor{lightgray}
           &    &     & L3 & 100 & 94 & 97 \\ \hline
Ch-pur(MP) & MU & 2024 & L1 & 14 & 32 & 19 \\ 
           &    &     & L2 & 13 & 35 & 19 \\ 
           \rowcolor{lightgray}
           &    &     & L3 & 95 & 91 & 93 \\ \hline
H-garh(RJ) & WH & 2024 & L1 & 46 & 66 & 54 \\ 
           &    &     & L2 & 46 & 73 & 56 \\ 
           \rowcolor{lightgray}
           &    &     & L3 & 90 & 86 & 88 \\ \hline
H-garh(RJ) & MU & 2024 & L1 & 53 & 22 & 31 \\ 
           &    &     & L2 & 53 & 25 & 34 \\ 
           \rowcolor{lightgray}
           &    &     & L3 & 71 & 70 & 70 \\ \hline
Tonk(RJ)  & WH & 2024 & L1 & 53 & 17 & 26 \\ 
          &    &     & L2 & 53 & 19 & 28 \\ 
          \rowcolor{lightgray}
          &    &     & L3 & 72 & 55 & 62 \\ \hline
Tonk(RJ)  & MU & 2024 & L1 & 52 & 62 & 56 \\ 
          &    &     & L2 & 52 & 69 & 59 \\ 
          \rowcolor{lightgray}
          &    &     & L3 & 91 & 80 & 85 \\ \hline

\end{tabular}
}
\caption{Metrics for different districts, crops, and levels of GT cleaning. (Ch-pur is Chhatarpur and H-garh is Hanumangarh)}
\label{tab:performance_metrics}
\end{table}
.
% \usepackage{booktabs}
%

%
\vspace{-1cm}
\bibliographystyle{IEEEbib}
\bibliography{GT_cleaning_InGARSS24}

\begin{thebibliography}{1}

\bibitem{ghosh2008problems}
Nilabja Ghosh and SS~Yadav,
\newblock ``Problems and prospects of crop insurance: reviewing agricultural risk and nais in india,''
\newblock {\em Institute of Economic Growth of Delhi Enclave North Campus, Delhi: http://www. iegindia. org/ardl/2008\_Crop\% 20Insurance}, vol. 20, 2008.

\bibitem{benami2021uniting}
Elinor Benami, Zhenong Jin, Michael~R Carter, Aniruddha Ghosh, Robert~J Hijmans, Andrew Hobbs, Benson Kenduiywo, and David~B Lobell,
\newblock ``Uniting remote sensing, crop modelling and economics for agricultural risk management,''
\newblock {\em Nature Reviews Earth \& Environment}, vol. 2, no. 2, pp. 140--159, 2021.

\bibitem{lin2022early}
Chenxi Lin, Liheng Zhong, Xiao-Peng Song, Jinwei Dong, David~B Lobell, and Zhenong Jin,
\newblock ``Early-and in-season crop type mapping without current-year ground truth: Generating labels from historical information via a topology-based approach,''
\newblock {\em Remote Sensing of Environment}, vol. 274, pp. 112994, 2022.

\bibitem{kumar2017statistical}
Pradeep Kumar, Rajendra Prasad, Arti Choudhary, Varun~Narayan Mishra, Dileep~Kumar Gupta, and Prashant~K Srivastava,
\newblock ``A statistical significance of differences in classification accuracy of crop types using different classification algorithms,''
\newblock {\em Geocarto International}, vol. 32, no. 2, pp. 206--224, 2017.

\bibitem{joshi2023remote}
Abhasha Joshi, Biswajeet Pradhan, Shilpa Gite, and Subrata Chakraborty,
\newblock ``Remote-sensing data and deep-learning techniques in crop mapping and yield prediction: A systematic review,''
\newblock {\em Remote Sensing}, vol. 15, no. 8, pp. 2014, 2023.

\bibitem{scikit-learn}
F.~Pedregosa, G.~Varoquaux, A.~Gramfort, V.~Michel, B.~Thirion, O.~Grisel, M.~Blondel, P.~Prettenhofer, R.~Weiss, V.~Dubourg, J.~Vanderplas, A.~Passos, D.~Cournapeau, M.~Brucher, M.~Perrot, and E.~Duchesnay,
\newblock ``Scikit-learn: Machine learning in python,''
\newblock {\em Journal of Machine Learning Research}, vol. 12, pp. 2825--2830, 2011.

\bibitem{lightgbm}
Qi~Meng GuolinKe, Thomas Finley, Taifeng Wang, Wei Chen, Weidong Ma, Qiwei Ye, and Tie-Yan Liu,
\newblock ``Lightgbm: A highly efficient gradient boosting decision tree,''
\newblock {\em Adv. Neural Inf. Process. Syst}, vol. 30, pp. 52, 2017.

\end{thebibliography}
\newpage

\end{document}